\documentclass[letterpaper, 10 pt, conference]{ieeeconf}  

\usepackage{amsmath}
\usepackage{graphicx}
\usepackage{multirow}
\usepackage{multicol}
\usepackage{array}
\usepackage{makecell}
\usepackage{indentfirst}
\usepackage{url}
\usepackage{amssymb}
\usepackage{lipsum}
\usepackage{placeins}
\usepackage{booktabs, caption}
\usepackage{array}
\usepackage{booktabs}
\usepackage{multirow}
\usepackage{wrapfig}
\usepackage{titlesec}
\usepackage{comment}
\usepackage{gensymb}
 \usepackage{longtable}
\usepackage{anyfontsize}
\usepackage{soul}
\usepackage{changepage,threeparttable}
\usepackage{tabularx}
\usepackage[normalem]{ulem}
\usepackage[usestackEOL]{stackengine}
\usepackage{subcaption}
\usepackage{listings}
\usepackage{float}
\usepackage{siunitx}

\IEEEoverridecommandlockouts                              

\title{\LARGE \bf
From Orbit Matching to Docking: A Hierarchical Reinforcement-Learning Approach to Multiphase Rendezvous Trajectory Design
}

\author{Mariko Storey-Matsutani$^{1}$, Anatoly Zlotnik$^{2}$, Lee Holguin$^{2}$, Zigfried Hampel-Arias$^{2}$, Richard Linares$^{3}$
\thanks{$^{1}$Mariko Storey-Matsutani is with the Department of Aeronautics and Astronautics, Massachusetts Institute of Technology,
        Cambridge, MA 02139, USA. Email: 
        {\tt\small m\_storey@mit.edu}}%
\thanks{$^{2}$Anatoly Zlotnik, Lee Holguin, and Zigfried Hampel-Arias are with the Los Alamos National Laboratory, Los Alamos, NM, USA. Emails:
        {\tt\small {azlotnik | holguin | zhampel}@lanl.gov}}%
\thanks{$^{1}$Richard Linares is an Associate Professor in the Department of Aeronautics and Astronautics, Massachusetts Institute of Technology,
        Cambridge, MA 02139, USA. Email: 
        {\tt\small linaresr@mit.edu}}%
}

\begin{document}

\maketitle
\thispagestyle{empty}
\pagestyle{empty}

\begin{abstract}

As the complexity of space missions increases, autonomous spacecraft guidance must accommodate transitions between mission phases with distinct dynamics, objectives, and constraints. Rendezvous and proximity operations (RPO) are particularly challenging because they span multiple operating regimes, from orbital-scale matching and far-range approach to close-range maneuvering and terminal docking. Existing reinforcement-learning approaches generally address only a subset of this sequence or rely on manually defined switching conditions between controllers. This study presents a hierarchical reinforcement-learning framework for autonomous, three-degree-of-freedom RPO that spans the complete trajectory from orbit matching to docking. Five specialized policies are trained for orbit matching, far-range approach, mid-range approach, close-range approach, and docking. A learned high-level manager policy then selects among these phase-specific controllers based on the evolving spacecraft state. The hierarchical approach is evaluated over the full end-to-end rendezvous sequence and compared with a fixed sequential-invocation heuristic to assess trajectory performance and transition behavior. Training and inference metrics are also analyzed for the individual phase policies and the complete hierarchical system.

\end{abstract}


\section{INTRODUCTION}
Rendezvous and proximity operations (RPO) are an essential component of many space missions, encompassing spacecraft docking, on-orbit servicing, active debris removal, and more. RPO trajectories require a chaser spacecraft to move from an initially separated orbit to a controlled relative state near a target. The full maneuver can be conceptually divided into distinct phases, such as orbit matching, approach, and docking. Each phase has different guidance objectives (eg. minimum fuel, minimum time), control time scales, and safety constraints. This close coordination between two or more spacecraft often requires the trajectory to be continually revised to account for state estimation uncertainty, spacecraft anomalies, or changing mission constraints during execution. Thus, autonomous RPO algorithms must be able to quickly regenerate updated control actions. 

Reinforcement learning (RL) has been studied as an alternative to traditional spacecraft guidance and control because a trained policy can produce state-dependent actions without repeatedly solving a computationally expensive optimization problem during execution. Early work applied RL to spacecraft rendezvous in cluttered environments by combining learned guidance with obstacle-avoidance methods \cite{broida_spacecraft_2019}. Subsequent studies extended these ideas to continuous-thrust rendezvous guidance \cite{wang_autonomous_2020}, experimental demonstrations of deep-RL controllers \cite{hovell_deep_2021}, and terminal rendezvous under stochastic disturbances \cite{federici_deep_2021}. More recent work has considered six-degree-of-freedom docking with rotating targets \cite{oestreich_autonomous_2021}, proximity maneuvering with collision avoidance \cite{qu_spacecraft_2022}, integrated rendezvous-and-docking trajectory planning \cite{chen_planning_2023}, and far-range rendezvous with safety and observability considerations \cite{wijayatunga_robust_2025}. Learning-based decomposition has also been explored through multiphase model-based RL for close-range approach and docking \cite{aborizk_multiphase_2024} and hierarchical RL for coordinating translational rendezvous guidance with attitude control \cite{tammam_hierarchical_2025}. Standardized environments have further enabled comparisons among RL algorithms for close-range rendezvous, docking, and attitude-control tasks \cite{supreeth_spacegym_2026}.

Collectively, these studies demonstrate that RL can learn effective policies for individual RPO tasks. However, existing literature generally treats orbit matching, far-range rendezvous, close-range proximity operations, and docking as separate control problems. A complete mission spans substantially different spatial and temporal scales, dynamical regimes, operational constraints, and control objectives. Furthermore, many existing RL formulations employ linearized Clohessy--Wiltshire dynamics \cite{federici_deep_2021,aborizk_multiphase_2024,supreeth_spacegym_2026, kaczmarek_autonomous_2023}, which are well suited to relative motion near a circular reference orbit but are not generally appropriate when the chaser and target begin in distant orbital states. Far-range orbit matching instead requires accounting for the nonlinear orbital motion of both vehicles before transitioning to the relative-motion descriptions commonly used during proximity operations. Consequently, policies designed for close-range motion cannot be directly extended to the complete orbital-scale maneuver. Existing multiphase approaches have considered long-range rendezvous through terminal docking, but do not explicitly include the orbit-matching maneuvers required to bring spacecraft from distinct orbital states into a common rendezvous regime \cite{aborizk_multiphase_2024,kaczmarek_autonomous_2023,gill_multi-phase_2026}. To the authors' knowledge, no prior RL formulation integrates orbit matching, far-range rendezvous, intermediate proximity operations, terminal approach, and docking within a learned control architecture.

This paper presents a hierarchical RL approach for autonomous, three-degree-of-freedom RPO that addresses this full multiphase problem.  First, we train a set of specialized policies, each dedicated to a distinct phase of the rendezvous trajectory: orbit matching, far-range approach, mid-range approach, close-range approach, and docking. Then, we train a high-level RL manager policy that learns to transition between the single-phase models to form a complete end-to-end maneuver. Rather than relying on hard-coded switching logic, the high-level policy learns directly from experience when to invoke each phase, producing a single autonomous system that spans the full multiphase mission from orbit matching to docking. Unlike prior hierarchical RL work for RPO that focuses on close-range operations under linearized dynamics, our approach encompasses the complete trajectory range using nonlinear dynamics modeled in the Earth-Centric Inertial (ECI) frame. We implement this system using Proximal Policy Optimization (PPO) with neural network policies, providing a computationally efficient solution for real-time onboard execution. We evaluate this hierarchical RL policy and compare against a fixed sequential-invocation heuristic to demonstrate reliable performance across the full end-to-end maneuver. We then perform an analysis of the computational requirements for each single-phase and multiphase model during training and inference.

The remainder of this paper is organized as follows. We first present the RPO problem formulation, including the spacecraft state and control variables, equations of motion, problem constraints,  and rendezvous phase definitions. The RL approach is then described, including the reward component formulations, single-phase RL model architectures and the hierarchical managers. This is followed by the single-phase and multiphase trajectory results, as well as an analysis of each model's computational requirements. Finally, we conclude with an overview of the main findings, limitations, and areas for future work.

\section{RENDEZVOUS AND PROXIMITY OPERATIONS}

\subsection{Orbital Dynamics Formulation}
An RL model framework that can learn both far-range and close range maneuvering requires a dynamics formulation that is appropriate for both regimes, without linearized simplifications such as the Hill-Clohessy-Wiltshire (HCW) equations \cite{clohessy_terminal_1960}. We define the orbital dynamics as an orbit propagation in ECI coordinates, similar to the formulation given by Wijayatunga et al. \cite{wijayatunga_robust_2025}. We omit the modeling of orbital perturbations and assume that spacecraft motion is dominated by Earth's gravitational force and the spacecraft's own actuation, but make no further assumptions about the orbital parameters or relative motion in the dynamics formulation. Table \ref{tab:notation} defines the relevant notation used throughout the dynamics formulation and the remainder of the paper.

    \begin{table}[h]
    \vspace{2ex}
  \centering
  \caption{Notation and Variable Definitions}
  \label{tab:notation}
  \resizebox{.98\linewidth}{!}{%
  \begin{tabular}{cc}
    \toprule
    Notation/Variable     & Definition  \\
    \midrule
    $r$ & Position \\
    $v$ & Velocity \\
    $F$ & Thrust force (control) \\
    $m$ & Mass \\
    $t$ & Time \\
    $[B, C, T]$ & [Body, Chaser, Target] \\
    $[\mathcal{I}, \mathcal{L}]$ & [Inertial (Earth-Centered), Target LVLH (Hill)] Frame\\
    $\mathcal{R}_{\mathcal{L}\to\mathcal{I}}$ & Rotation from LVLH to inertial frame \\
    $\mu$ & Earth Gravitational Parameter \\
    $\omega_\mathcal{L}$ & LVLH frame angular velocity \\
    $\rho$ & Relative range, $\|r_{rel}^{\mathcal{L}}\|$ \\
    $\theta_{\mathrm{appch}}$ & Approach angle (total cone angle) \\
    $\hat{a}_{\mathrm{appch}}^{\mathcal{L}}$ & Approach cone axis, $[1, 0, 0]^\top$ \\
    $k$ & Control step index \\
    $k_f$ & Final control step \\
    $N$ & Success condition hold duration (steps) \\
    \bottomrule
  \end{tabular}}
\end{table}

The same gravitational two-body dynamics apply to both the chaser and the target, so the state and dynamics below are written for a general body $B \in \{C,T\}$. The chaser is actuated, while the target is assumed to be a non-maneuvering body. Control is commanded in the target-centered Local-Vertical-Local-Horizontal (LVLH) frame $\mathcal{L}$, otherwise known as the Hill frame, defined from the target's instantaneous state \cite{clohessy_terminal_1960}.

\textbf{State Vector:}
\begin{equation}
x_B
=
\begin{bmatrix}
r_B^{\mathcal{I}} \\
v_B^{\mathcal{I}}
\end{bmatrix}
\end{equation}

\textbf{Control Vector:}
\begin{equation}
u_B
=
F_B^{\mathcal{L}}
\in \mathbb{R}^{3}
\end{equation}

\textbf{Coordinate Frame Rotation:}
\begin{align}
\hat{e}_{R}^{\mathcal{I}}
&=
\frac{r_T^{\mathcal{I}}}
{\left\|r_T^{\mathcal{I}}\right\|_2}
\\[3pt]
\hat{e}_{N}^{\mathcal{I}}
&=
\frac{h_T^{\mathcal{I}}}
{\left\|h_T^{\mathcal{I}}\right\|_2},  \text{ where  } h_T^{\mathcal{I}}
=
r_T^{\mathcal{I}} \times v_T^{\mathcal{I}}
\\[3pt]
\hat{e}_{T}^{\mathcal{I}}
&=
\hat{e}_{N}^{\mathcal{I}}
\times
\hat{e}_{R}^{\mathcal{I}}
\end{align}

\begin{equation}
\mathcal{R}_{\mathcal{I}\to\mathcal{L}}
=
\begin{bmatrix}
\left(\hat{e}_{R}^{\mathcal{I}}\right)^\top \\[2pt]
\left(\hat{e}_{T}^{\mathcal{I}}\right)^\top \\[2pt]
\left(\hat{e}_{N}^{\mathcal{I}}\right)^\top
\end{bmatrix},
\qquad
\mathcal{R}_{\mathcal{L}\to\mathcal{I}}
=
\mathcal{R}_{\mathcal{I}\to\mathcal{L}}^\top.
\end{equation}

\textbf{Dynamics:}
\begin{equation}
\dot{x}_B
=
\begin{bmatrix}
v_B^{\mathcal{I}} \\[5pt]
-\dfrac{\mu}
{\left\|r_B^{\mathcal{I}}\right\|_2^3}
r_B^{\mathcal{I}}
+
\mathcal{R}_{\mathcal{L}\to\mathcal{I}}
\dfrac{F_B^{\mathcal{L}}}{m_B}
\end{bmatrix}
\end{equation}

\textbf{Relative Motion:}
\begin{align}
r_{\mathrm{rel}}^{\mathcal{L}}
&=
\mathcal{R}_{\mathcal{I}\to\mathcal{L}}
\left(
r_C^{\mathcal{I}} - r_T^{\mathcal{I}}
\right),
\\[3pt]
v_{\mathrm{rel}}^{\mathcal{L}}
&=
\mathcal{R}_{\mathcal{I}\to\mathcal{L}}
\left(
v_C^{\mathcal{I}} - v_T^{\mathcal{I}}
\right)
-
\omega_{\mathcal{L}/\mathcal{I}}^{\mathcal{L}}
\times
r_{\mathrm{rel}}^{\mathcal{L}},
\\[3pt]
\text{where}&
\qquad
\omega_{\mathcal{L}/\mathcal{I}}^{\mathcal{L}}
=
\begin{bmatrix}
0 \\[2pt]
0 \\[2pt]
\dfrac{\left\|h_T^{\mathcal{I}}\right\|_2}
{\left\|r_T^{\mathcal{I}}\right\|_2^2}
\end{bmatrix}\notag
\end{align}

\begin{equation}
\theta_{\mathrm{appch}}
=
\arccos\left(
\frac{\hat{a}_{\mathrm{appch}}^{\mathcal{L}} \cdot r_{\mathrm{rel}}^{\mathcal{L}}}
{\left\|r_{\mathrm{rel}}^{\mathcal{L}}\right\|_2}
\right)
\end{equation}

\textbf{Constraints:}
\begin{align}
\text{Idle Target: }
& \quad F_T^{\mathcal{L}} = 0\\[3pt]
\text{Thrust Bound: }
&\quad
\left\|
\frac{F_C^{\mathcal{L}}}{m_C}
\right\|_\infty
\le
a_{\max},
\\[3pt]
\text{Range Bound: }
&\quad
\left\|
r_{\mathrm{rel}}^{\mathcal{L}}
\right\|_2
\le
\rho_{\max},
\\[3pt]
\text{Delta-V Budget: }
&\quad
\sum_{k=0}^{k_f-1}
\Delta v_k
\le
\Delta v_{\max},
\\[2pt]
&\quad
\text{where}
\qquad
\Delta v_k
=
\left\|
\frac{F_{C,k}^{\mathcal{L}}}{m_C}
\right\|_2
\Delta t,
\notag \\[3pt]
\text{Timestep Bound: }
&\quad
0 \le k \le k_{\max}.
\end{align}

\textbf{Success Condition:} \\
\paragraph*{$ \forall k \in \{k_f-N+1,\ldots,k_f\}:$}
\begin{align}
\left\|r_{\mathrm{rel},k}^{\mathcal{L}}\right\|_2 &\le \rho_{\mathrm{tol}}, \\[3pt]
\left\|v_{\mathrm{rel},k}^{\mathcal{L}}\right\|_2 &\le v_{\mathrm{tol}}, \\[3pt]
\theta_{\mathrm{appch},k} &\le \theta_{\mathrm{tol}}.
\end{align}

This formulation provides a single dynamically consistent model across all mission phases, avoiding the range and orbit-geometry limitations of local linearized models while remaining simple enough for large-scale RL training. In Section \ref{sec:problem def}, we describe the initial and terminal conditions of each trajectory phase, in addition to constraint bound values on the acceleration, range, $\Delta v$, and time used when training the RL models. The approach angle success condition is enforced only for the close-range, docking, and hierarchical manager models.

\subsection{Rendezvous Trajectory Phase Definitions}

Rather than learning a single RL policy for the full RPO control protocol from orbit matching to docking, we divide the full RPO problem into five distinct phases: orbit matching, far-range approach, mid-range approach, close-range approach, and docking. This is advantageous because a feasible reward formulation for orbit matching differs vastly from a feasible reward formulation for docking behavior. Table \ref{tab:rpo_phases} outlines the initial and terminal conditions used to define each phase of the RPO trajectory, as well as the corresponding modeling parameters. At very far initial separations, the initial condition becomes easier to characterize in terms of orbital elements rather than relative position and velocity.

\label{sec:problem def}

\begin{table}[H]
\centering
\caption{Rendezvous Approach Phases}
\label{tab:rpo_phases}
\resizebox{\linewidth}{!}{%
\begin{tabular}{lccl}
\toprule
\textbf{Phase} & \textbf{Initial Condition} & \textbf{Terminal Condition} & \textbf{Model Parameters} \\ 
\midrule
\multirow{5}{*}{Orbit Match} & Phase: [-90.0, 90.0]$^\circ$  & $\rho_{\mathrm{tol}} = 10{,}000$ m & $\Delta v_{\max} = 10{,}000$ m/s \\[1pt]
  & SMA: [-20k, 20k] m & $v_{\mathrm{tol}} = 50$ m/s & $\Delta v_{\mathrm{k,max}} = 1200$ m/s \\[1pt]
  & Ecc.: [0.0, 0.01] & & $\rho_{\max} = 25{,}000$ km \\[1pt]
  & Inc.: [-45.0, 45.0]$^\circ$ & & $k_{\max} =$ 180 \\[1pt]
  & RAAN: [-45.0, 45.0]$^\circ$ & & \\[2pt]
\midrule
\multirow{4}{*}{Far Range} & $\|r_{\mathrm{rel},0}^{\mathcal{L}}\|$: [3000, 10000] m & $\rho_{\mathrm{tol}} = 1000$ m & $\Delta v_{\max} = 180$ m/s \\[2pt]
  & $\|v_{\mathrm{rel},0}^{\mathcal{L}}\|$: [0.0, 1.0] m/s & $v_{\mathrm{tol}} = 0.5$ m/s & $\rho_{\max} = 20{,}000$ m \\[2pt]
  & & & $a_{\max} = 0.05$ m/s$^2$ \\[1pt]
  & & & $k_{\max} =$ 500 \\[3pt]
\midrule
\multirow{4}{*}{Mid Range} & $\|r_{\mathrm{rel},0}^{\mathcal{L}}\|$: [800, 1200] m & $\rho_{\mathrm{tol}} = 100$ m & $\Delta v_{\max} = 70.7$ m/s \\[2pt]
  & $\|v_{\mathrm{rel},0}^{\mathcal{L}}\|$: [0.0, 0.3] m/s & $v_{\mathrm{tol}} = 0.3$ m/s & $\rho_{\max} = 3{,}000$ m \\[2pt]
  & & & $a_{\max} = 0.05$ m/s$^2$ \\[1pt]
  & & & $k_{\max} =$ 900 \\[3pt]
\midrule
\multirow{4}{*}{Close Range} & $\|r_{\mathrm{rel},0}^{\mathcal{L}}\|$: [80.0, 120.0] m& $\rho_{\mathrm{tol}} = 10$ m & $\Delta v_{\max} = 50$ m/s \\[2pt]
  & $\|v_{\mathrm{rel},0}^{\mathcal{L}}\|$: [0.0, 0.3] m/s& $v_{\mathrm{tol}} = 0.2$ m/s & $\rho_{\max} = 300$ m \\[2pt]
  & $\theta_{\mathrm{appch},0} = 60.0^\circ$ & $\theta_{\mathrm{tol}} = 30^\circ$ & $a_{\max} = 0.05$ m/s$^2$ \\[1pt]
  & & & $k_{\max} =$ 900 \\[3pt]
\midrule
\multirow{4}{*}{Docking} & $\|r_{\mathrm{rel},0}^{\mathcal{L}}\|$: [8.0, 12.0] m& $\rho_{\mathrm{tol}} = 1$ m & $\Delta v_{\max} = 20$ m/s \\[2pt]
  & $\|v_{\mathrm{rel},0}^{\mathcal{L}}\|$: [0.0, 0.2] m/s& $v_{\mathrm{tol}} = 0.1$ m/s & $\rho_{\max} = 150$ m \\[2pt]
  & $\theta_{\mathrm{appch},0} = 60.0^\circ$ & $\theta_{\mathrm{tol}} = 30^\circ$ & $a_{\max} = 0.05$ m/s$^2$ \\[1pt]
  & & & $k_{\max} =$ 900 \\
\bottomrule
\end{tabular}}
\end{table}

\section{Reinforcement Learning Architecture}
Reinforcement Learning (RL) is a machine learning framework that learns a reward-maximizing policy through environment interaction. RL formulates problems as Markov Decision Processes (MDP): 

\begin{equation} \mathcal{M}=\left(\mathcal{X},\mathcal{U},p,\phi,\chi_0,\gamma\right), \end{equation}

where $\mathcal{X}$ and $\mathcal{U}$ denote the state and action spaces, $p(x_{t+1}\mid u_t,x_t)$ is the transition distribution, $\phi$ is the reward function, $\chi_0$ is the initial-state distribution, and $\gamma\in[0,1]$ is the discount factor. At each time step, a policy $\pi_\theta(u_t\mid x_t)$ parameterized by $\theta$ selects an action from the current state and the environment produces a corresponding reward and next state \cite{sutton2018reinforcement, puterman1994markov}.

For a task $\mathcal{M}$, the expected return of policy $\pi_\theta$ is

\begin{equation} J_{\mathcal{M}}(\theta) = \mathbb{E}_{\tau \sim p_{\theta}(\tau \mid \mathcal{M})} \left[ \sum_{t=0}^{H-1} \gamma^t \phi_t \right], \end{equation}

where $\tau$ is a trajectory generated by the policy in task $\mathcal{M}$ and the corresponding RL objective is
\begin{equation} \arg\max_{\theta} J_{\mathcal{M}}(\theta). \end{equation}

Proximal policy optimization (PPO) is an  actor-critic method used for RL training. Actor-critic algorithms use a policy network (actor) $\pi_{\theta}$ to select actions and a value network (critic) $V_{\psi}$ to estimate expected returns, with the critic providing advantage estimates that reduce variance in policy gradient updates. PPO is a common choice in using RL for control design where simulation is relatively inexpensive and the action space is typically continuous. Its on-policy nature requires many fresh samples for training but provides stable policy updates through its clipped objective function, which prevents destructively large policy changes that could be catastrophic in continuous control tasks \cite{schulman2017proximal, konda2000actor}.

\subsection{Reward Design}
\label{sec:reward}
Reward design is a critical element of RL model training. For our RPO problem, we define the reward components given in Table \ref{tab:reward_components}, which we divide conceptually into four groups. The first group consists of potential-based progress terms, including relative position and orbital-element progress. These rewards measure the change in error between timesteps, so the reward reflects the controller’s net improvement rather than absolute error. This is especially useful at long range, where a per-timestep error penalty can unintentionally incentivize fast episode termination. The second group contains continuous regulation terms, which shape how the vehicle moves rather than where it is. Velocity and range-rate penalties regulate approach speed, while the coast incentive discourages unnecessary thrusting. The third group includes precision terms like a station keeping reward and hold-streak bonus that rewards remaining within position and velocity tolerances, while the exponential error and position/velocity penalty terms provide a stronger incentive for highly accurate convergence. The remaining terms provide general trajectory efficiency and outcome signals. The $\Delta v$ penalty discourages excessive thrusting, the time penalty discourages unnecessarily long solutions, the terminal condition captures success or failure, and the skill no-op penalty prevents the hierarchical manager from repeatedly selecting a skill whose subgoal has already been satisfied.

\begin{table}[h]
\vspace{2ex}
\centering
\caption{Reward Component Definitions}
\label{tab:reward_components}
\begin{tabular}{p{1.3cm}|p{6.5cm}}
\textbf{Reward} & \textbf{Definition} \\
\hline
Position Penalty & $\phi = \begin{cases} r_{rel}/10 & r_{rel} < r_{\text{prox}} \\ 0 & \text{otherwise} \end{cases}$ \\
\hline
Velocity Penalty& Constant: $\phi = -||\mathbf{v}_{\text{rel}}||$ \\
& Distance-Scaled: $\phi = -\dfrac{||\mathbf{v}_{\text{rel}}||}{\max(r_{rel}, r_{\min})}$ \\
\hline
Progress & $\phi = \dfrac{\epsilon_{\text{prev}} - \epsilon_{\text{curr}}}{s}$, where \\
 & $\epsilon \in$ \{altitude difference $|h_{\text{tgt}} - h|$, \qquad (s=8M); \\
 & \qquad eccentricity difference $|e_{\text{tgt}} - e|$,\quad (s=0.005);\\
 & \qquad inclination difference $|i_{\text{tgt}} - i|$,\qquad (s=0.01);\\
 & \qquad relative position $||\mathbf{r}_{\text{chaser}} - \mathbf{r}_{\text{tgt}}||$, \\
 & \qquad true anomaly difference $|\theta_{\text{tgt}} - \theta|$, \quad (s=2.0);\\
 & \qquad plane alignment $1 - \hat{\mathbf{n}}_{\text{chaser}} \cdot \hat{\mathbf{n}}_{\text{tgt}}$, \qquad(s=5E-4)\} \\
\hline
Range-Rate Control & $\phi = \begin{cases} -(1E-5)(v_{rel} - v_{\max})^2 & v_{rel} < -v_{\max} \\ 0.1(1 - r_{rel}/r_{\text{act}}) & r_{rel} < r_{\text{act}} \\ -(5E-5)v_{rel} & v_{rel} > 0 \\ 0 & \text{otherwise} \end{cases}$ \\
 & $v_{\max} = v_{\max,\text{near}}$ if $r < r_{\text{gate}}$, else $v_{\max,\text{far}}$ \\
\hline
Coast Incentive & $\phi = \begin{cases} -||\mathbf{a}_{\text{cmd}}|| & r_{rel} < r_{\text{prox}} \\ 0 & \text{otherwise} \end{cases}$ \\
\hline
Safety \;\;Penalty & $\phi = \begin{cases} -(3E-8)(h_{\text{warn}} - h) & h < h_{\text{warn}} \\ 0 & \text{otherwise} \end{cases}$ \\
\hline
Station \;\; Keeping & $\phi = \begin{cases} 10 & r_{rel} < r_{\text{tol}}, \\ &v_{\text{rel}} < v_{\text{tol}} \\ -(1E-8)[(\frac{r_{rel}-r_{SK}}{r_{\text{tol}}})^2 + (\frac{v_{\text{rel}}}{v_{\text{tol}}})^2] & \text{otherwise} \end{cases}$ \\
\hline
Hold Bonus & $\phi = n_{\text{hold}} \cdot b$ \\
\hline
Approach Cone & $\phi = \begin{cases} -(\theta_{appch} - \theta_{tol})^2/s & \theta_{appch} > \theta_{\text{tol}}, \text{ and}\\
& r_{rel} < r_{appch} \\ 0 & \text{otherwise} \end{cases}$ \\
\hline
Exponential Error & $\phi = \exp\!\left(-\left\lVert \boldsymbol{\epsilon} \right\rVert\right)$ (or $\exp(-\lVert\boldsymbol{\epsilon}\rVert^2)$), \text{where} \\
& $\boldsymbol{\epsilon} \in \left\{\dfrac{h - h_{\text{tgt}}}{3000},\ \dfrac{i - i_{\text{tgt}}}{0.02},\ \dfrac{1-\hat{\mathbf{n}}\cdot\hat{\mathbf{n}}_{\text{tgt}}}{0.0005},\ \dfrac{\lVert \mathbf{h}_{\text{tgt}} - \mathbf{h}\rVert}{0.05 \cdot\lVert \mathbf{h}_{\text{tgt}}\rVert} \right\}$ \\
\hline
Delta-v Penalty & Linear: $\phi = -\Delta v$, \\
 & Quadratic: $\phi = -\Delta v^2$ \\
 \hline
Time Penalty & $\phi = -1$ \\
\hline
Skill No-Op Penalty & $\phi = \begin{cases} -1 & \Delta t \le t_{\min} \text{ and } \Delta r < \Delta r_{\min} \\ 0 & \text{otherwise} \end{cases}$ \\
\hline
Terminal \;\;Condition & $\phi = \begin{cases} 100 & \text{if success,} \\ -100 & \text{otherwise} \end{cases}$ \\
\hline
\end{tabular}
\vspace{-2ex}
\end{table}

\subsection{Single-Phase RL Controllers}
Using the reward components defined in Section \ref{sec:reward}, we first train a set of single-phase controllers, each specialized for a distinct portion of the overall rendezvous and docking problem. The far-range and mid-range approach models rely on reward terms that consider relative position and velocity, and the close-range and docking models employ an additional approach-cone corridor to constrain the final approach direction. In contrast, the orbit-matching model considers semi-major axis (SMA), eccentricity, inclination, orbital-plane, and true-anomaly errors that are better suited to impulsive $\Delta v$ actions. The orbit-matching controller is trained over a three-stage curriculum that progressively widens the inclination and right ascension (RAAN) offset ($5^\circ \rightarrow 20^\circ \rightarrow 45^\circ$) and scales the per-step $\Delta v$ budget (400 m/s $\rightarrow$ 600 m/s $\rightarrow$ 1200 m/s) while keeping the reward formulation constant. Table \ref{tab:reward_vals} lists the reward weights for all controllers, and Table \ref{tab:reward_components} defines reward components.

\begin{table}[h]
\vspace{1ex}
\centering
\caption{Reward Weights Across All Controllers}
\label{tab:reward_vals}
\setlength{\tabcolsep}{2pt}
\resizebox{\linewidth}{!}{%
\begin{tabular}{l||c|c|c|c|c||c}
 & \multicolumn{5}{c||}{\textbf{Single-Phase Models}} & \\
\hline
 & & & & & & \\
\textbf{Parameter} & \textbf{\shortstack{Orbit\\Match}} & \textbf{\shortstack{Far\\Range}} & \textbf{\shortstack{Mid\\Range}} & \textbf{\shortstack{Close\\Range}} & \textbf{Dock} & \textbf{\shortstack{Multi\\Phase}} \\
\hline
\textit{\textbf{Reward Weights}} & & & & & & \\
Position Penalty & - & - & - & -1.0 & -1.0 & - \\
Velocity Penalty &   &   &   &      &      &  \\
$\quad$ Distance-Scaled & - & 1.0 & 1.0 & - & - & 1.0 \\
$\quad$ Constant & - & - & - & 1.0 & 1.0 & - \\
\shortstack{Position Progress\\\scriptsize(Scale Value)} & - & \shortstack{1.0\\\scriptsize(s=50)} & \shortstack{1.0\\\scriptsize(s=50)} & - & - & \shortstack{1.0\\\scriptsize(s=50k)} \\
\shortstack{Approach Cone\\\scriptsize(Scale Value)} & - & - & - & \shortstack{1.0\\\scriptsize(s=1.0)} & \shortstack{1.0\\\scriptsize(s=0.1)} & - \\
Altitude Progress & 1.0 & - & - & - & - & - \\
Eccentricity Progress & 10.0 & - & - & - & - & - \\
Inclination Progress & 10.0 & - & - & - & - & - \\
True Anomaly Progress & 10.0 & - & - & - & - & - \\
Plane Progress & 10.0 & - & - & - & - & - \\
Range Rate Control & 1.0 & - & - & - & - & - \\
Coast Incentive & 0.1 & - & - & - & - & - \\
Safety Penalty  & 1.0 & - & - & - & - & - \\
Station-Keeping  & 1.0 & - & - & - & - & - \\
Hold Streak Bonus & 1.0 & - & - & - & - & - \\
Exponential Error & 10.0 & - & - & - & - & - \\
Delta-v Penalty &  &  &  &  &  &  \\
$\quad$ Linear & - & 1.0 & 1.0 & 1.0 & 1.0 & 1.0 \\
$\quad$ Quadratic & 0.0012 & - & - & - & - & - \\
Time Penalty & 0.0 & 0.1 & 0.1 & - & - & 0.1 \\
Skill No-Op Penalty & - & - & - & - & - & 5.0 \\
Terminal Success & 1.0 & 1.0 & 1.0 & 1.0 & 1.0 & 1.0 \\
Terminal Failure & 0.5 & 1.0 & 1.0 & 1.0 & 5.0 & 1.0 \\
\end{tabular}}
\vspace{-2ex}
\end{table}

All controllers use the same two-body relative-motion environment, PPO for training, and compact multi-layer perceptron (MLP) policy architecture given in Table \ref {tab:training}. Only the initial condition distribution, success tolerances, $\Delta v$ and range limits, and reward weights are changed across stages. Fixing the simulator, learning algorithm, and policy capacity allows reward designs to be compared without confounding due to differences in model size or optimization method.

\subsection{Multiphase Hierarchical RL Manager}
We train a hierarchical reinforcement learning manager to select among the trained single-phase controllers. The manager operates at a coarser temporal resolution than the low-level controllers and is trained using PPO with a discrete action space corresponding to the five available skills. After a skill is selected, its frozen policy controls the spacecraft for up to 950 low-level environment steps before the manager selects again. The 950 step-cap is chosen such that it comfortably exceeds each single-phase model's own mean episode length. To simplify the large-range operating regime, the orbit-matching controller is selected whenever the chaser-target separation exceeds 100km. At smaller separations, the manager may select any of the five controllers, including an earlier-phase controller when the current state requires recovery or correction. A macro-step can also terminate before reaching the 950-step limit when the active controller satisfies its local handoff condition. This allows control to return to the manager as soon as the current phase objective is achieved and limits operation of each controller outside its training distribution.

The manager is trained using a dense shaping reward constructed from similar general reward components used for the single-phase controllers. Each episode's initial conditions include offsets in phase angle, semi-major axis (SMA), inclination (Inc), right ascension (RAAN), and eccentricity (Ecc), consistent with the orbit-matching training distribution. Observations and rewards are normalized online to stabilize value-function learning against the large reward-scale swings between a normal in-distribution macro-step and an out-of-distribution skill invocation. Since each phase transition requires the simulation of multiple low-level steps under a frozen controller, hierarchical training has substantially lower effective throughput than single-phase training. Table \ref{tab:reward_vals} lists the reward parameters used for the hierarchical multiphase model, while the corresponding reward-component definitions are provided in Table \ref{tab:reward_components}.

\begin{table}[h]
  \centering
  \caption{RL Model Parameters}
  \label{tab:training}
  \resizebox{\linewidth}{!}{%
  \begin{tabular}{lccccc|c}
    \toprule
    \textbf{Parameter} & \textbf{\shortstack{Orbit\\Match}} & \textbf{\shortstack{Far\\Range}} & \textbf{\shortstack{Mid\\Range}} & \textbf{\shortstack{Close\\Range}} & \textbf{Docking} & \textbf{\shortstack{Multi\\Phase}}\\ [2pt]
    \midrule
    Network Size & \multicolumn{5}{c|}{[128, 128]} & [128, 128] \\[2pt]
    Activation & \multicolumn{5}{c|}{tanh} & tanh\\[2pt]
    Learning Rate & \multicolumn{5}{c|}{0.0003} & 0.0003\\[2pt]
    Batch size & \multicolumn{5}{c|}{256}& 256\\[2pt]
    Discount  & 0.998 & 0.99 & 0.99 & 0.99 & 0.99 & 0.99\\[2pt]
    GAE $\lambda$ & \multicolumn{5}{c|}{0.95}& 0.95\\[2pt]
    Entropy Coef. & 0.01 & 0.0 & 0.0 & 0.0 & 0.0 & 0.01\\[2pt]
    Target KL & 0.03 & None & None & 0.03 & None & None\\
    \bottomrule
  \end{tabular}}
\end{table}

All six controllers -- five single-phase and one multiphase -- were trained independently with Stable-Baselines3's PPO implementation. Table \ref{tab:training} outlines the values for all training parameters in this study that differ from the default value in Stable-Baselines3 \cite{ran_stable-baselines3_nodate}.

\section{RESULTS}
\subsection{Single-Phase Model Results}
Before composing the single-phase models into the multiphase hierarchical policy, we evaluate each of the five single-phase controllers in isolation. Each policy is evaluated over $N=1000$ held-out episodes with a deterministic policy and a fixed evaluation seed. Table~\ref{tab:phase-summary} summarizes each model's success rate, final range, final velocity, and total $\Delta v$ statistics at episode termination.

\begin{table}[H]
  \centering
  \caption{Single-phase policy Evaluation
    ($N=1000$ episodes each, deterministic policy, held-out seed, all except success rate are median values). }
  \label{tab:phase-summary}
  \resizebox{\linewidth}{!}{%
  \begin{tabular}{lccccc}
    \toprule
    Phase & Success & Min. Range & Final Range  & Final Vel.  &  $\Delta v$  \\
     & Rate & (m) & (m) & (m/s) & (m/s)  \\
    \midrule
    Orbit Match    & 97.0\%  & 5380.10 & 5383.52 & 2.005 & 5660.64    \\
    Far Range      & 100.0\% & 249.96  & 269.97  & 0.478 & 26.83    \\
    Mid Range      & 100.0\% & 53.52   & 64.48   & 0.296 & 8.99    \\
    Close Range    & 100.0\% & 9.80    & 9.95    & 0.023 & 2.66     \\
    Docking        & 100.0\% & 0.98    & 0.98    & 0.009 & 0.25     \\
    \bottomrule
  \end{tabular}}
\end{table}

Four of five phases attain a perfect $100.0\%$ success rate over their held-out evaluation set, with the exception of the orbit matching policy, which reaches $97.0\%$ success. This validates that each single-phase policy is individually reliable within its own operating regime. As a performance benchmark, the best-performing SpaceGym baseline reports success rates of (82\%) for docking and (62\%) for rendezvous \cite{supreeth_spacegym_2026}. Although the SpaceGym tasks differ in exact initial and terminal conditions, the comparison highlights the strong performance of each single-phase model in this study. The median $\Delta v$ decreases monotonically with the shrinking size of the operating range, from $5660.64$\,m/s for orbit matching to $0.25$\,m/s for docking. This is consistent with the two-order-of-magnitude decrease in starting range and the correspondingly limited burn authority each closer-range phase is allowed.  

\begin{figure}[h]
  \centering
  \includegraphics[width=\linewidth]{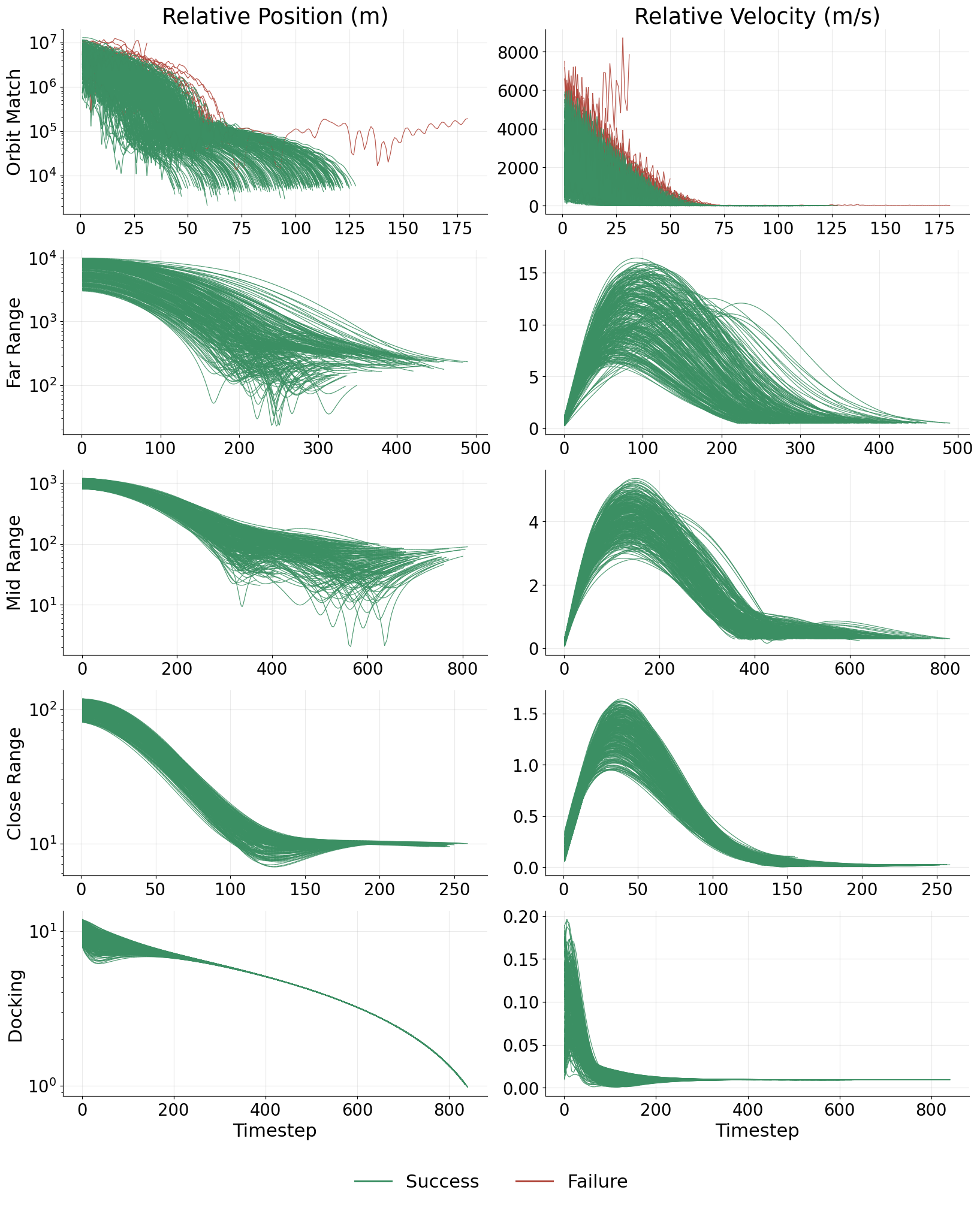}
  \caption{Relative Position and Velocity Magnitude }
  \label{fig:flat-traj}
  \vspace{-2ex}
\end{figure}

Comparing each model's median final range against its own position tolerance reveals each policy's success margin. For example, the far-range model converges to only $27.0\%$ of its $1000$\,m tolerance, leaving comfortable margins. By contrast, the close-range and docking models converge to $99.5\%$ and $97.6\%$ of their $10$\,m and $1$\,m tolerances, respectively. We attribute this to the additional 30 degree approach-angle success condition imposed on these proximity phases. The combined approach cone and proximity constraints likely consume most of the policy's available Delta-v budget, leaving little incentive to converge to a tighter range after the success condition is met.

\subsection{Multiphase Hierarchical Model Results}
\label{sec:hier-results}

We evaluate the trained hierarchical manager over N=1000 held-out episodes with a fixed seed, sampled from the same initial condition distribution as the single-phase orbit match model. The learned manager achieves 99.4\% success, with terminal accuracy within tolerance on both position and velocity. The median result is 0.976m and 0.00930m/s for the terminal position and velocity respectively, against the 1m and 0.1m/s tolerance requirement.

\begin{figure}[h]
    \centering
    \includegraphics[width=0.49\linewidth]{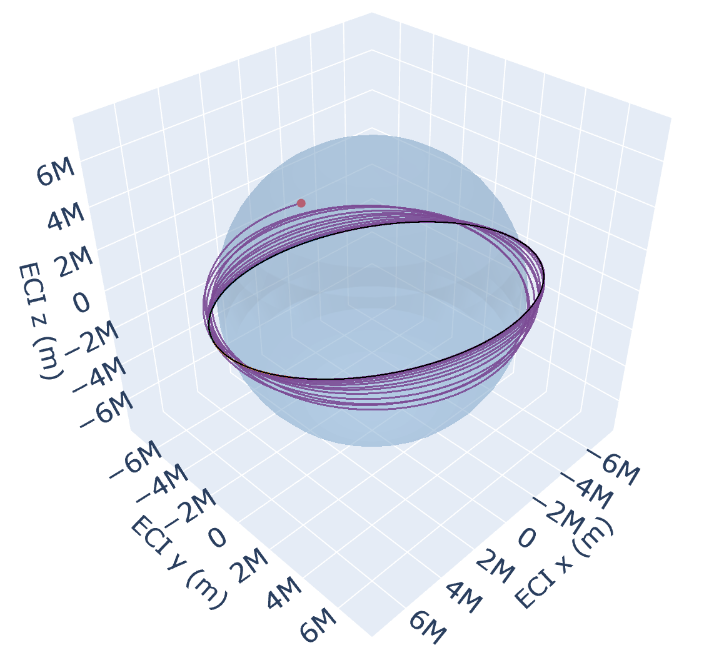}
    \includegraphics[width=0.48\linewidth]{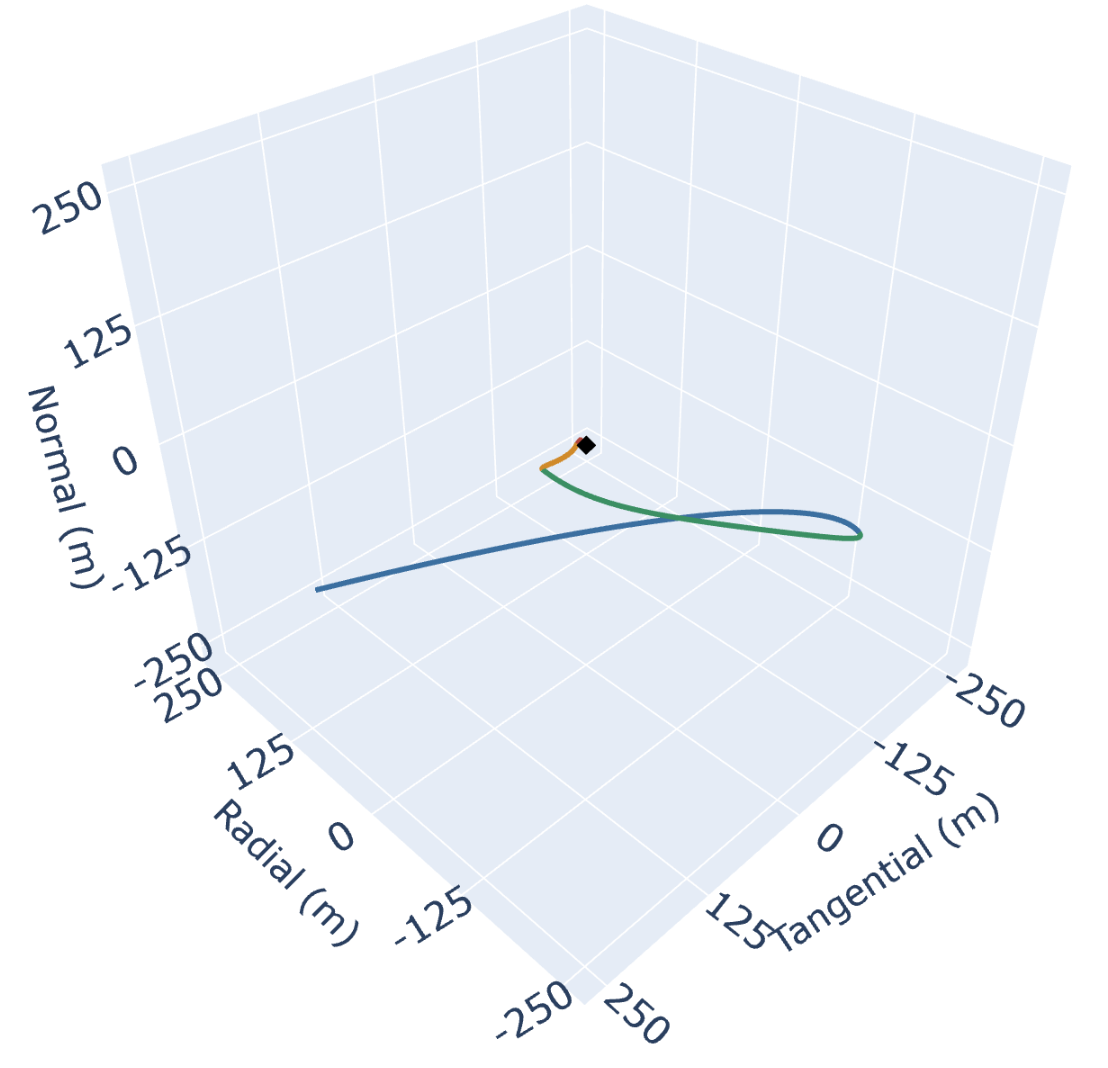}
    \includegraphics[width=\linewidth]{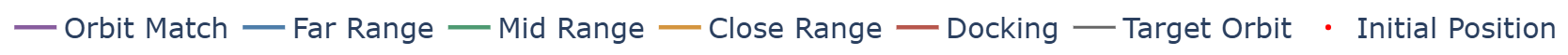}
    \caption{Trajectory Visualization from Learned Manager (Left: Earth-Centric, Right: Target-Centric)}
    \label{fig:trajs}
\end{figure}

To contextualize these results, we compare the learned manager against a heuristic baseline that runs the same five single-phase models exactly once each, in a fixed order determined by each model's own termination tolerance. Without any shared hierarchical environment, each model executes in its own independent environment under its own full dynamics and tolerances, with no shared timestep cap and no shared Delta-v budget across models. Instead, only the terminal physical state is carried forward to the next model as its initial state. As seen in Table \ref{tab:manager-vs-baseline}, the heuristic achieves 97.0\% success at a median $\Delta v$ of 5684.86m/s, while the learned manager reaches a higher 99.4\% success rate at a comparable 5706.86m/s median $\Delta v$. 

\begin{table}[h]
  \centering
  \caption{Comparison of learned manager vs. heuristic baseline performance metrics with N=1000 (Format: Median (Mean $\pm$ Std))}
  \label{tab:manager-vs-baseline}
  \resizebox{0.98\linewidth}{!}{%
  \begin{tabular}{lcc}
    \toprule
    Metric & Learned manager & Sequential Heuristic \\
    \midrule
    Success rate                      & 99.4\% & 97.0\% \\ [2pt]
    Final Range (m)        & 0.976            & 0.977 \\
                           & ($3.76e4 \pm 5.13e5$) & ($5.25e4 \pm 6.37e5$) \\[2pt]
    Final Velocity (m/s)          & 0.00930            & 0.00930 \\
                                  & ($66.6 \pm 947$) & ($40.4 \pm 525$) \\[2pt]
    $\Delta v$ (m/s)          & 5706.86            & 5684.86 \\
                              & ($5.67e3 \pm 2.85e3$) & ($5.55e3 \pm 2.41e3$) \\[2pt]
    \bottomrule
  \end{tabular}}
\end{table}

The two approaches converge to nearly identical terminal margins: a median final range of 0.976m and velocity 0.00930m/s for the learned manager, against 0.977m and 0.00930m/s for the heuristic. Both results are within the 1m and 0.1m/s tolerance requirements. This behavior is consistent with each single-phase model handing off exactly at its own success threshold rather than continuing to refine the approach once nominally within tolerance. We note that the mean--std statistics are dominated by the small number of failed episodes in the 1000-episode sample (6 for the learned manager, 30 for the heuristic) and should thus be taken in context with the median results.

\begin{figure}[h]
    \centering
    \includegraphics[width=\linewidth]{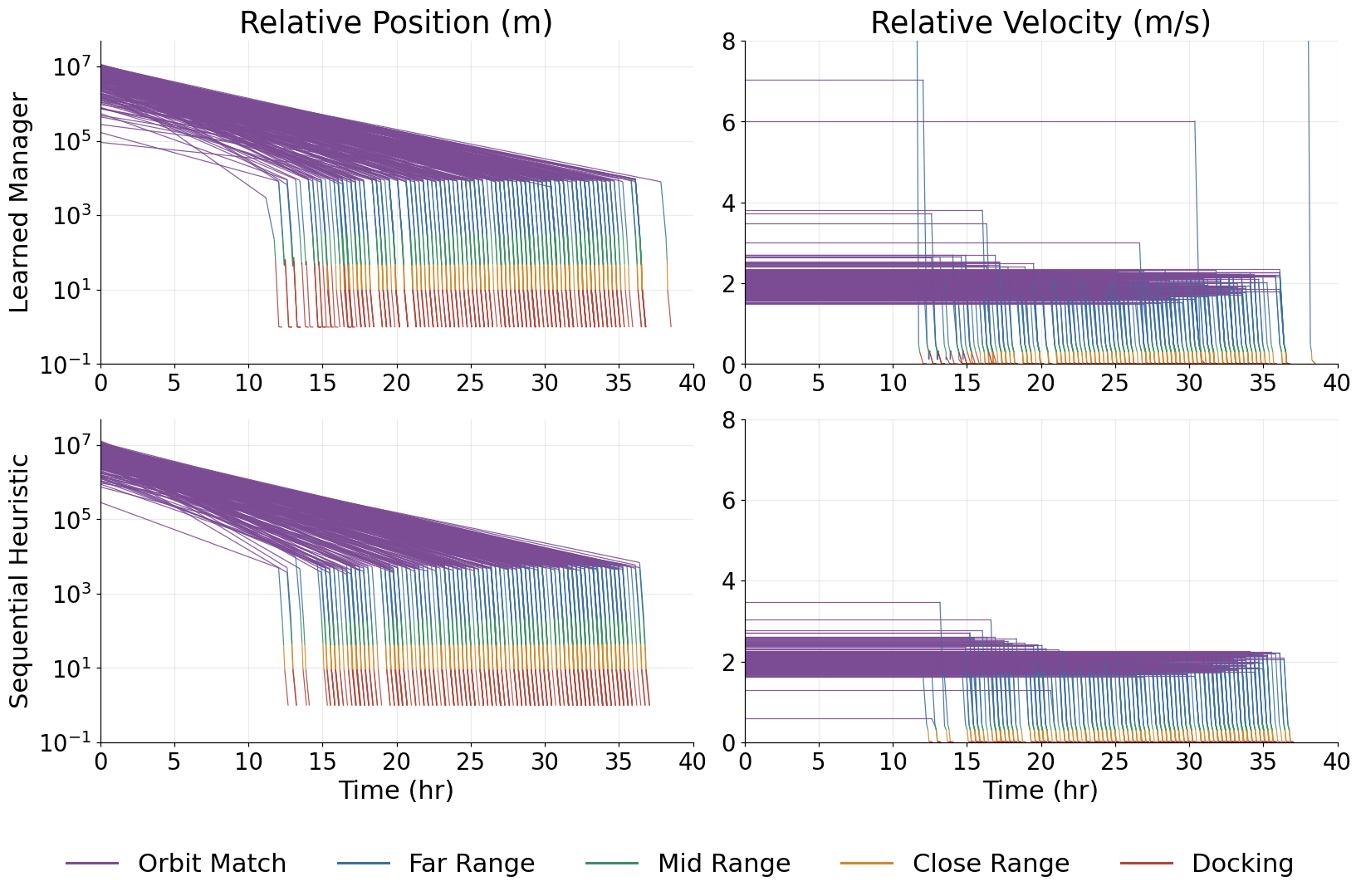}
    \caption{Comparison of learned manager vs. heuristic baseline for relative position and velocity through each trajectory. Only successful trajectories are displayed.}
    \label{fig:manager_comp}
\end{figure}

Next, we investigate how each manager's time is distributed across the five phases, as presented in Table \ref{tab:phase-time}. The learned manager spends a majority of each trajectory in the orbit-matching phase (97.141\%), with the far-range, mid-range, close-range, and docking models together accounting for under 3\%. The heuristic shows a similar pattern, with 97.491\% of time spent in orbit-matching. This is consistent with the physical intuition behind the trajectory design, where far-range orbital-plane matching is a slow, coast-dominated maneuver, while the near-field approach and docking sequence is relatively fast, although it requires many more individual control decisions. The learned manager's mean trajectory duration is 26.6 hours, similar to the heuristic's 26.6 hours.

\begin{table}[h]
  \centering
  \caption{Time breakdown of learned manager and heuristic baseline by phase (percentages and hours reported as mean $\pm$ std across episodes)}
  \label{tab:phase-time}
 \resizebox{0.98\linewidth}{!}{%
 \begin{tabular}{lcc}
    \toprule
    Phase & Learned Manager & Heuristic \\
    \midrule
    Orbit Match   & 97.141\% $\pm$ 1.052\% & 97.491\% $\pm$ 0.803\% \\
                  & ($25.8836 \pm 7.4155$ h) & ($25.9421 \pm 6.3255$ h) \\[2pt]
    Far Range     & 1.161\% $\pm$ 0.380\% & 0.970\% $\pm$ 0.303\% \\
                  & ($0.2868 \pm 0.0275$ h) & ($0.2456 \pm 0.0448$ h) \\[2pt]
    Mid Range     & 0.484\% $\pm$ 0.154\% & 0.375\% $\pm$ 0.121\% \\
                  & ($0.1194 \pm 0.0109$ h) & ($0.0945 \pm 0.0174$ h) \\[2pt]
    Close Range   & 0.197\% $\pm$ 0.069\% & 0.265\% $\pm$ 0.087\% \\
                  & ($0.0507 \pm 0.0125$ h) & ($0.0665 \pm 0.0117$ h) \\[2pt]
    Docking       & 1.018\% $\pm$ 0.561\% & 0.899\% $\pm$ 0.295\% \\
                  & ($0.2437 \pm 0.0543$ h) & ($0.2263 \pm 0.0398$ h) \\
    \bottomrule
  \end{tabular}}
\end{table}

\subsection{Computational Resource Analysis}
The computational analysis was conducted on an Intel Core Ultra 7 268V CPU (8 cores/8 threads) with 32 GB of RAM, using single-threaded execution without parallelization or GPU acceleration.
Training was parallelized across 8 concurrent processes, one per physical core, using CPU-only execution.

Table~\ref{tab:training-compute} reports the training compute cost of each policy. The five phase-specific skills range from 0.49M to 4.01M total timesteps seen during training, with per-skill training times of up to 1.56 hours. The hierarchical manager required substantially more training time since each of its high-level decisions requires a roll-out of the single-phase policies.

\begin{table}[h]
\vspace{2ex}
\centering
\caption{Training metrics per model}
\label{tab:training-compute}
  \resizebox{0.98\linewidth}{!}{%
\begin{tabular}{lccc}
\toprule
 & Timesteps & Training & Throughput\\
Model & & Time (h) & (fps) \\
\midrule

Orbit Match &    &  &   \\
\;\;\;\; Stage 1 & 491,520   & 0.06 & 2,296  \\
\;\;\;\; Stage 2 & 430,848   & 0.05 & 2,256  \\
\;\;\;\; Stage 3 & 2,179,072 & 0.29 & 2,072    \\
\;\;\;\; Total & 3,101,440 & 0.40 & 2,129   \\
Far Range        & 491,520   & 0.09 & 1,510    \\
Mid Range        & 688,128   & 0.17 & 1,135    \\
Close Range      & 4,014,080 & 1.18 & 942  \\
Docking          & 3,031,040 & 1.56 & 539 \\
Multiphase       & 199,968   & 3.02 & 18.4 \\
\bottomrule
\end{tabular}}
\vspace{2ex}
\end{table}

We characterize each policy's computational cost using four metrics:
\begin{enumerate}
    \item Parameter count: the number of trainable weights in the network
    \item Checkpoint size: the on-disk storage footprint of the saved model
    \item Inference latency: the mean wall-clock time required to compute a single action from an observation
    \item Control-step margin: the ratio of the simulated control step duration to the mean inference latency
\end{enumerate}

All six learned policies -- the five phase-specific controllers and the hierarchical manager -- use the same two-layer actor-critic network, differing only in observation and action dimensions. Single-sample inference latency is therefore similar across models, ranging from 0.093 to 0.117 ms. At this scale, the measured latency is largely dominated by Stable-Baselines3 and Gymnasium call overhead rather than network computation \cite{ran_stable-baselines3_nodate, towers2024gymnasium}. For comparison, Bashnick and Ulrich report mean online solution times that are nearly four orders of magnitude larger at 0.34-4.82 s for a three-dimensional rendezvous problem, although differences in hardware and implementation preclude a direct speedup comparison \cite{Bashnick2023FastMPC}. Table \ref{tab:footprint-latency} reports the model size, inference latency, and control step margin of each model. This real-time margin ranges from ($8.53\times10^{3}$) for the close-range skill to ($9.35\times10^{6}$) for the orbit matching model, whose 1031.8s macro-step interval makes inference cost negligible relative to the control timescale. The hierarchical manager itself achieves a comparable ($8.62\times10^{3}$) margin evaluated against the same 1 Hz reference cadence.

\begin{table}[H]
\centering
\caption{Model size and inference latency (N = 1{,}000)}
\label{tab:footprint-latency}

  \resizebox{\linewidth}{!}{%
\begin{tabular}{lcccc}
\toprule
 & Params & Checkpoint& Latency & Control Step\\
Model &  & (KB) & (ms) & Margin\\
\midrule
Docking          & 35,847 & 462 & $0.095 \pm 0.032$ & $1.05\times10^{4}$    \\
Close Range      & 35,847 & 464 & $0.117 \pm 0.072$ & $8.53\times10^{3}$  \\
Mid Range        & 35,847 & 464 & $0.093 \pm 0.037$ & $1.08\times10^{4}$ \\
Far Range        & 35,847 & 464 & $0.096 \pm 0.053$ & $3.75\times10^{4}$  \\
Orbit Matching   & 36,871 & 476 & $0.110 \pm 0.053$ & $9.35\times10^{6}$  \\
Multiphase & 37,126 & 484 & $0.116 \pm 0.062$ & $8.62\times10^{3}$ \\
\bottomrule
\end{tabular}}
\end{table}

 As presented in Table~\ref{tab:hier-rollout}, the physics/environment operations account for 64.74\% of the learned manager's episode time and 63.86\% of the heuristic baseline's episode time. Inference contributes only 0.21\% of wall-clock time under the learned hierarchical manager, while the heuristic baseline avoids any manager inference given its deterministic hand-off policy. Over $N=1{,}000$ episodes with identical initialization, both approaches complete in under one second per episode on average. The learned manager's mean episode time of 0.961s is modestly higher than the heuristic's 0.917s, but this gap is an artifact of the cases on which the heuristic fails. On the episodes where the two disagree on success, the heuristic terminates in $\sim$0.058s by failing fast and halting the skill chain, whereas the learned manager expends $\sim$0.963s to complete the docking maneuver. On the episodes both controllers solve, their per-episode costs are comparable, and the learned manager exhibits lower run-to-run variance ($\pm$0.152s vs. $\pm$0.221s), reflecting more consistent skill sequencing. Accounting for task completion, the learned manager's 99.4\% success rate corresponds to 0.967s of expected computation per successful docking, compared to 0.945s for the sequential baseline's 97.0\% success rate.

\begin{table}[H]
\centering
\vspace{-2ex}
\footnotesize
\caption{Computational breakdown for the learned and heuristic hierarchical managers (N = 1{,}000 paired episodes)}
\label{tab:hier-rollout}
\begin{tabular}{lcc}
\toprule
Metrics & Learned & Heuristic \\
\midrule
Success rate                          & 0.994            & 0.970 \\
Mean episode wall-clock (s)           & $0.961 \pm 0.152$ & $0.917 \pm 0.221$ \\
Median episode wall-clock (s)         & 0.932  & 0.905 \\
Expected wall-clock per success (s)   & 0.967  & 0.945 \\
\midrule
\textit{\textbf{Computational Breakdown}} & & \\
Manager Inference (\%)                & 0.21           & - \\
Single-Phase Inference (\%)                  & 35.05           & 36.14 \\
Physics/Environment Simulation (\%)              & 64.74           & 63.86 \\
\bottomrule
\end{tabular}
\vspace{-1ex}
\end{table}

\section{CONCLUSION}

This work presents a hierarchical reinforcement learning framework for autonomous rendezvous and proximity operations that successfully integrates the complete mission sequence from orbit matching to terminal docking within a learned control architecture. By training five specialized phase-specific policies and a high-level manager that learns state-dependent phase transitions, the approach achieves 99.4\% success rate over 1,000 held-out evaluation episodes while maintaining comparable Delta-v efficiency (median $\Delta v$ of 5706.86 m/s) to a fixed sequential baseline (97.0\% success, 5684.86 m/s). The use of nonlinear orbital dynamics in the ECI frame enables the framework to handle large initial separations that cannot be adequately addressed by linearized relative-motion models. The hierarchical structure allows each phase to be optimized independently with phase-appropriate reward formulations before integration into the complete system. Computational analysis demonstrates that the learned policies are suitable for real-time onboard execution, with inference latencies under 0.12 milliseconds and control-step margins exceeding $10^3$. 

Several limitations remain: the current formulation considers only three-degree-of-freedom  translational dynamics without perturbations, assumes an idle target, and does not explicitly model sensor noise or actuation uncertainties that would be present in operational systems. Additionally, further reward tuning could improve model training times and trajectory behavior.

\addtolength{\textheight}{-12cm}   

\section*{ACKNOWLEDGMENT}
We gratefully acknowledge Andrew Rogers for his insight on RPO trajectory design, as well as Christopher Koh and Kshitija Taywade for their early contributions to code development. This project was supported by the LDRD program project 20240415ER ``Initial Orbital Determination using Adaptive Optimization'' at Los Alamos National Laboratory. Research conducted at Los Alamos National Laboratory is done under the auspices of the National Nuclear Security Administration of the U.S. Department of Energy under Contract No. 89233218CNA000001. Report LA-UR-26-27724. 



\bibliographystyle{unsrt}
\bibliography{sources}

@article{towers2024gymnasium,
  title={Gymnasium: A standard interface for reinforcement learning environments},
  author={Towers, Mark and Kwiatkowski, Ariel and Balis, John and De Cola, Gianluca and Deleu, Tristan and Goul{\~a}o, Manuel and Andreas, Kallinteris and others},
  journal={Advances in Neural Information Processing Systems},
  volume={38},
  year={2026}
}

@book{sutton2018reinforcement,
  title={Reinforcement learning: An introduction},
  author={Sutton, Richard S. and Barto, Andrew G.},
  year={2018},
  edition={2nd},
  publisher={MIT Press},
  address={Cambridge, MA}
}

@book{puterman1994markov,
  title={Markov decision processes: Discrete stochastic dynamic programming},
  author={Puterman, Martin L.},
  year={1994},
  publisher={John Wiley \& Sons},
  address={New York, NY}
}

@article{schulman2017proximal,
  title={Proximal policy optimization algorithms},
  author={Schulman, John and Wolski, Filip and Dhariwal, Prafulla and Radford, Alec and Klimov, Oleg},
  journal={arXiv preprint arXiv:1707.06347},
  year={2017}
}

@inproceedings{konda2000actor,
  title={Actor-critic algorithms},
  author={Konda, Vijay R. and Tsitsiklis, John N.},
  booktitle={Advances in Neural Information Processing Systems},
  volume={12},
  pages={1008--1014},
  year={2000}
}

@inproceedings{kaczmarek_autonomous_2023,
  author    = {Kaczmarek, Jeremy J. and Zagaris, Constantinos},
  title     = {Autonomous Multi-Phase Rendezvous, Proximity Operations,
               and Docking via Model Predictive Control},
  booktitle = {IEEE Aerospace Conference},
  year      = {2023},
  pages     = {1--10},
  doi       = {10.1109/AERO55745.2023.10115983}
}

@inproceedings{chen_planning_2023,
  author    = {Chen, Vincent and Phillips, Sean A. and Copp, David A.},
  title     = {Planning Autonomous Spacecraft Rendezvous and Docking
               Trajectories via Reinforcement Learning},
  booktitle = {Proceedings of the 45th Annual AAS Guidance,
               Navigation and Control Conference},
  year      = {2023},
  address   = {Breckenridge, Colorado},
  note      = {AAS Paper 23-026}
}

@article{tammam_hierarchical_2025,
  author  = {Tammam, Abdulla and Aouf, Nabil},
  title   = {Hierarchical Deep Reinforcement Learning for {CubeSat}
             Guidance and Control},
  journal = {Control Engineering Practice},
  volume  = {156},
  pages   = {106213},
  year    = {2025},
  doi     = {10.1016/j.conengprac.2024.106213}
}

@article{wijayatunga_robust_2025,
  author  = {Wijayatunga, Minduli Charithma and Armellin, Roberto
             and Holt, Harry},
  title   = {Robust Trajectory Design and Guidance for Far-Range
             Rendezvous Using Reinforcement Learning with Safety
             and Observability Considerations},
  journal = {Aerospace Science and Technology},
  volume  = {159},
  pages   = {109996},
  year    = {2025},
  doi     = {10.1016/j.ast.2025.109996}
}

@article{Bashnick2023FastMPC,
  author  = {Bashnick, Courtney and Ulrich, Steve},
  title   = {Fast Model Predictive Control for Spacecraft Rendezvous
             and Docking with Obstacle Avoidance},
  journal = {Journal of Guidance, Control, and Dynamics},
  volume  = {46},
  number  = {5},
  pages   = {998--1007},
  year    = {2023},
  doi     = {10.2514/1.G007314}
}

@article{clohessy_terminal_1960,
  author  = {Clohessy, W. H. and Wiltshire, R. S.},
  title   = {Terminal Guidance System for Satellite Rendezvous},
  journal = {Journal of the Aerospace Sciences},
  year    = {1960},
  volume  = {27},
  number  = {9},
  pages   = {653--658},
  month   = sep,
  doi     = {10.2514/8.8704}
}

@article{qu_spacecraft_2022,
    title = {Spacecraft {Proximity} {Maneuvering} and {Rendezvous} {With} {Collision} {Avoidance} {Based} on {Reinforcement} {Learning}},
    volume = {58},
    issn = {1557-9603},
    url = {https://ieeexplore.ieee.org/abstract/document/9793683},
    doi = {10.1109/TAES.2022.3180271},
    number = {6},
    urldate = {2026-06-09},
    journal = {IEEE Transactions on Aerospace and Electronic Systems},
    author = {Qu, Qingyu and Liu, Kexin and Wang, Wei and Lü, Jinhu},
    month = dec,
    year = {2022},
    pages = {5823--5834},
}

@article{oestreich_autonomous_2021,
    title = {Autonomous {Six}-{Degree}-of-{Freedom} {Spacecraft} {Docking} with {Rotating} {Targets} via {Reinforcement} {Learning}},
    volume = {18},
    issn = {1940-3151},
    url = {https://doi.org/10.2514/1.I010914},
    doi = {10.2514/1.I010914},
    number = {7},
    urldate = {2026-06-09},
    journal = {Journal of Aerospace Information Systems},
    publisher = {American Institute of Aeronautics and Astronautics},
    author = {Oestreich, Charles E. and Linares, Richard and Gondhalekar, Ravi},
    year = {2021},
    pages = {417--428}
}

@inproceedings{supreeth_spacegym_2026,
    title = {{SpaceGym}: {A} {Gymnasium}-{Based} {Benchmark} for {Deep} {Reinforcement} {Learning} in {Spacecraft} {Proximity} {Operations}},
    shorttitle = {{SpaceGym}},
    url = {https://openaccess.thecvf.com/content/CVPR2026W/AI4Space/html/Supreeth_SpaceGym_A_Gymnasium-Based_Benchmark_for_Deep_Reinforcement_Learning_in_Spacecraft_CVPRW_2026_paper.html},
    language = {en},
    urldate = {2026-06-09},
    author = {Supreeth, Mysore and Mehta, Manish},
    year = {2026},
    pages = {10388--10396}
}

@article{aborizk_multiphase_2024,
    title = {Multiphase {Autonomous} {Docking} via {Model}-{Based} and {Hierarchical} {Reinforcement} {Learning}},
    volume = {61},
    issn = {0022-4650, 1533-6794},
    url = {https://arc.aiaa.org/doi/10.2514/1.A35683},
    doi = {10.2514/1.A35683},
    language = {en},
    number = {4},
    urldate = {2026-06-09},
    journal = {Journal of Spacecraft and Rockets},
    author = {Aborizk, Anthony and Fitz-Coy, Norman},
    month = jul,
    year = {2024},
    pages = {993--1005}
}

@inproceedings{broida_spacecraft_2019,
  title={{Spacecraft Rendezvous Guidance in Cluttered Environments via Reinforcement Learning}},
  author={Broida, Jacob and Linares, Richard},
  booktitle={29th AAS/AIAA Space Flight Mechanics Meeting},
  pages={1--15},
  year={2019}
}

@article{hovell_deep_2021,
    title = {Deep {Reinforcement} {Learning} for {Spacecraft} {Proximity} {Operations} {Guidance}},
    volume = {58},
    issn = {0022-4650},
    url = {https://arc.aiaa.org/doi/10.2514/1.A34838},
    doi = {10.2514/1.A34838},
    number = {2},
    urldate = {2026-07-17},
    journal = {Journal of Spacecraft and Rockets},
    publisher = {American Institute of Aeronautics and Astronautics},
    author = {Hovell, Kirk and Ulrich, Steve},
    month = mar,
    year = {2021},
    pages = {254--264},
}

@article{federici_deep_2021,
    title = {Deep {Learning} {Techniques} for {Autonomous} {Spacecraft} {Guidance} {During} {Proximity} {Operations}},
    volume = {58},
    issn = {0022-4650},
    url = {https://arc.aiaa.org/doi/10.2514/1.A35076},
    doi = {10.2514/1.A35076},
    number = {6},
    urldate = {2026-07-17},
    journal = {Journal of Spacecraft and Rockets},
    publisher = {American Institute of Aeronautics and Astronautics},
    author = {Federici, Lorenzo and Benedikter, Boris and Zavoli, Alessandro},
    month = nov,
    year = {2021},
    pages = {1774--1785},
}

@article{wang_autonomous_2020,
    title = {Autonomous {Rendezvous} {Guidance} via {Deep} {Reinforcement} {Learning}},
    issn = {1948-9447},
    url = {https://ieeexplore.ieee.org/document/9163988/},
    doi = {10.1109/CCDC49329.2020.9163988},
    urldate = {2026-07-17},
    journal = {2020 Chinese Control And Decision Conference (CCDC)},
    author = {Wang, Xinyu and Wang, Guohui and Chen, Yi and Xie, Yongfeng},
    month = aug,
    year = {2020},
    note = {ISSN: 1948-9447},
    pages = {1848--1853},
}

@article{ran_stable-baselines3_nodate,
  title={{Stable-Baselines3: Reliable Reinforcement Learning Implementations}},
  author={Raffin, Antonin and Hill, Ashley and Gleave, Adam and Kanervisto, Anssi and Ernestus, Maximilian and Dormann, Noah},
  journal={Journal of Machine Learning Research},
  volume={22},
  number={268},
  pages={1--8},
  year={2021}
}

@article{gill_multi-phase_2026,
    title = {Multi-{Phase} {Guidance} and {Control} for {Small} {Satellite} {Docking} via {Deep} {Reinforcement} {Learning}},
    url = {https://digitalcommons.usu.edu/smallsat/2026/all2026/140},
    journal = {Small Satellite Conference},
    author = {Gill, Tanner and Furlich, Greg and Ahmed, Nisar},
    month = aug,
    year = {2026},
}


\end{document}